\documentclass[journal]{IEEEtran}
\ifCLASSINFOpdf
\else
\fi
 \usepackage{threeparttable}
\usepackage{subfigure}
\usepackage{graphicx}
\usepackage{amsmath}
\usepackage{amssymb}
\usepackage{amsfonts}
\usepackage{mathtools}

\usepackage{stfloats}
\usepackage{float}
\usepackage{lipsum}
\usepackage{multirow}
\usepackage[pdftex,colorlinks,bookmarksopen,bookmarksnumbered,citecolor=red,urlcolor=red]{hyperref}

\usepackage{color} 
\usepackage[numbers,sort&compress]{natbib}
\restylefloat{figure}

\begin{document}
%
\title{\bf Post-Hurricane Damage Assessment Using Satellite Imagery and Geolocation Features\vspace{0.4em}}
%
%
%

\author{Quoc Dung Cao and Youngjun Choe\hspace{.2cm}\\
    Department of Industrial and Systems Engineering\\ University of Washington}

\maketitle

\begin{abstract}
Gaining timely and reliable situation awareness after hazard events such as a hurricane is crucial to emergency managers and first responders. One effective way to achieve that goal is through damage assessment. Recently, disaster researchers have been utilizing imagery captured through satellites or drones to quantify the number of flooded/damaged buildings. In this paper, we propose a mixed data approach, which leverages publicly available satellite imagery and geolocation features of the affected area to identify damaged buildings after a hurricane. The method demonstrated significant improvement from performing a similar task using only imagery features, based on a case study of Hurricane Harvey affecting Greater Houston area in 2017. This result opens door to a wide range of possibilities to unify the advancement in computer vision algorithms such as convolutional neural networks and traditional methods in damage assessment, for example, using flood depth or bare-earth topology. In this work, a creative choice of the geolocation features was made to provide extra information to the imagery features, but it is up to the users to decide which other features can be included to model the physical behavior of the events, depending on their domain knowledge and the type of disaster. The dataset curated in this work is made openly available (DOI: 10.17603/ds2-3cca-f398).
\end{abstract}



%
\IEEEpeerreviewmaketitle

\section{Introduction}
%
%
%
%
\IEEEPARstart{D}{amage} assessment after a hurricane landfall is increasingly important for emergency management as the hurricane intensity and frequency increase. The current practice of windshield survey, which relies on emergency response crews and volunteers to drive around the affected area is known to be costly and time-consuming. To speed up the process, several studies have been conducted to reduce data collection time or assist the visual inspection. One notable direction is using deep learning to detect whether a building is damaged or not after a hurricane event. In our previous work \cite{cao2020building}, we have shown that using satellite imagery and annotation labels from a crowdsourcing campaign can achieve state-of-the-art performance on classifying damaged buildings based on several metrics such as accuracy, precision-recall, and F1 score. Besides satellite imagery in pure Red-Green-Blue (RGB) band, other studies have explored the use of deep learning in flood risk assessment using IKONOS-2 multispectral and panchromatic imagery \cite{van2003segmentation}, or time series of Landsat-5/7  satellite imagery \cite{skakun2014flood}. Other works have also studied to perform building damage assessment through extracting building texture feature using synthetic aperture radar (SAR) data \cite{chen2019novel}, multitemporal very high resolution SAR imagery \cite{pirrone2020approach}, or multitemporal polarimetric SAR data \cite{chen2018urban}. In this work, we continue to use the high-resolution optical sensor RGB imagery since they are easier for emergency managers to interpret.

In fact, within the field of damage assessment, deep learning techniques have been showing promising results. A subset of deep learning models called convolutional neural network (CNN) have been applied to detect damage in concrete structures \cite{cha2017deep, cha2018autonomous, huang2018deep}, car damage \cite{patil2017deep, zhang2019crowdlearn}, or regional change detection after disaster events \cite{doshi2018satellite}. However, these methods rely heavily on the quantity and quality of the labelled dataset, which in some cases might be unavailable or noisy. Sometimes, performance can be capped in some large image dataset such as Imagenet \cite{cnn-imagenet} and improvement in performance is marginal, regardless of model architecture.  

Before the CNN era, there were established methods to assess flooding hazard risks, such as analyzing precipitation, catchment capacity, or river network analyses \cite{apel2004flood}, generating flood outlines and depth based on topological data \cite{hall2003methodology}, or simulating flood spreading \cite{gouldby2008methodology}. In the seismic risk analysis field, there are also probabilistic risk assessment \cite{ellingwood2001earthquake} or defining vulnerability indices for infrastructure systems \cite{pitilakis2006earthquake}. These methods are still extremely valuable even in these days as a natural hazard is a natural phenomenon that obeys physical rules.

The above observation inspires us to hypothesize that there could be a potential improvement to the post-hurricane damage assessment process if we can utilize multiple types of data. We propose to utilize the optical sensor satellite imagery and other geolocation features of the individual buildings in our damage assessment framework. This work will open up another possibility to understanding disaster damage. For example, for hurricanes, we can combine precipitation level, flooding resilience index, catchment capacity, elevation, or river networks with the imagery data to improve damage assessment and also understand which characteristics are more critical to the likelihood of damage. On the other hand, in seismic risk assessment, we can also incorporate the relative distance of the buildings/roads to the epicenter, Ritchter magnitude, ground shaking in the zone through various sensors, or seismic resilience index into the model, in addition to the aerial images. 

\begin{figure}[ht!]
{\centering
\includegraphics[width=3.4in]{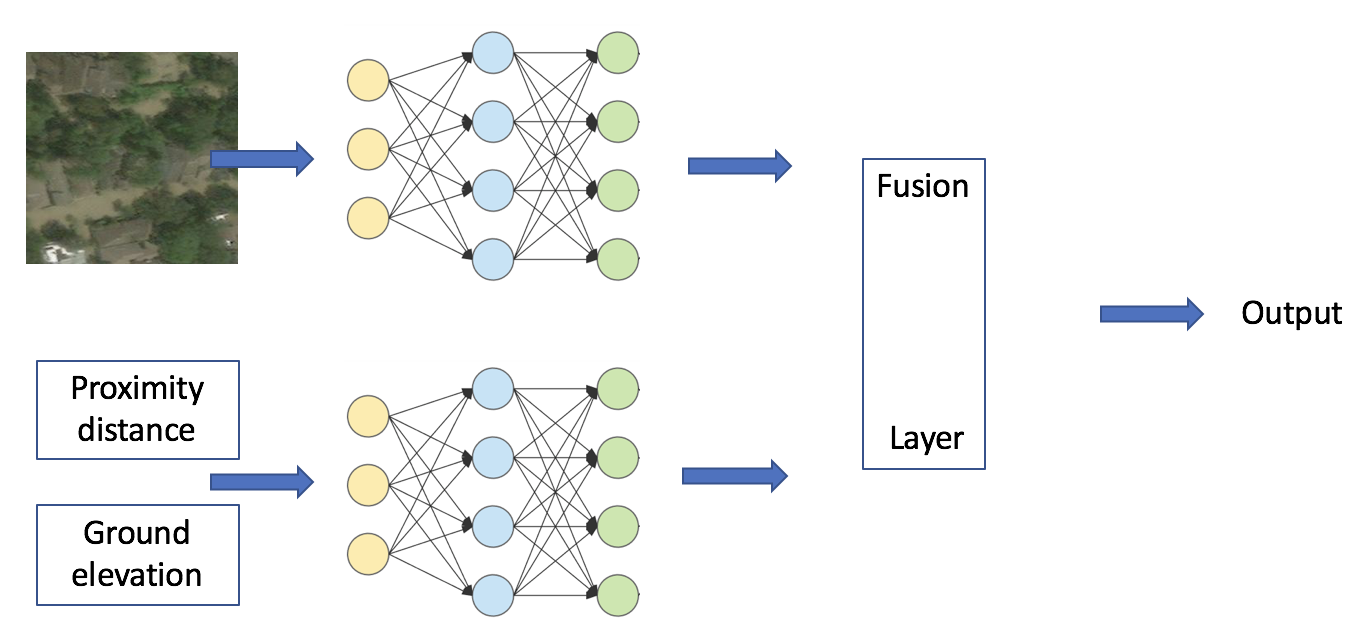} 
\centering\caption{\small{Workflow of the mixed data neural network model.}}
\label{fig:combined-intro}
}
\end{figure}

In this work, we present a mixed data approach to damage assessment by utilizing the satellite imagery and other geolocation features such as building elevation and proximity to water bodies (Figure~\ref{fig:combined-intro}) to improve the performance and generalizability of our previous work \cite{cao2020building}. Our contribution is two-fold. First, by considering mixed data, we can leverage more domain knowledge to understand disaster damage assessment better and boost its predictive performance. Second, as an improvement to our previous dataset in \cite{cao2020building}, we collect the \textit{`Undamaged Building'} labels and  \textit{`Flooded/Damaged'} labels from imagery of the same timestamp. Specifically, we manually build the \textit{`Undamaged Building'} labels from the undamaged region of the same imagery captured after the hurricane event. This dataset is more realistic and generalizable than the previous dataset using purely satellite imagery since it reflects the actual situation when we want to deploy this damage assessment framework in future events. The dataset is curated and made available on the DesignSafe Data Depot (DOI: 10.17603/ds2-3cca-f398) \cite{designsafe}.

The remaining of this paper is organized as follows. In Section~\ref{sec:background_multimodal}, we present a brief review of current literature on flood damage assessment and applications using mixed data. Section~\ref{sec:method_multimodal} describes our proposed methodology for dataset construction and model architecture. Details of the implementation and discussion of the results are presented in Section~\ref{sec:case_study_multimodal}. Finally, Section~\ref{sec:conclusion_multimodal} concludes this paper and draws some future research directions. 

\section{Background}\label{sec:background_multimodal}
\subsection{Relevant hurricane damage assessment practice}
In this section, we review some of the state-of-the-art methods in hurricane damage assessment. There has always been a close relationship between flooding properties, ground topography and damage quantification. Within the field of hydrology, the role of bare-earth topography elevation is so important to hydraulic modeling of water flows that the elevation estimation methods have been studied in various works such as \cite{yuan2019mapping, o2016multi}. In \cite{kwak2015rapid}, using moderate resolution imaging spectrometer (MODIS) time-series imagery, the crop damage extend at each map grid pixel (approximately 500m) is studied as a function of flood depth and flood duration. Similarly, a study by \cite{Mehrotra2015} attempts to classifies regions of pixels into water, vegetation, urban, and bare land at the coastal areas of Japan after an earthquake triggered a tsunami. In a separate study, relative frequency of flood inundation is shown to exhibit the same probabilistic distribution as relative water depth, which is characterized by bed elevation \cite{skakun2014flood}.

The above methods mostly utilize variants of SAR imagery and/or earth elevation from digital elevation model (DEM) in their work. SAR imagery has its own advantages in mapping surface features, or roughness pattern. However, it could be harder for laymen (e.g emergency managers and first responders) to interpret than optical sensor imagery. In addition, there are fewer satellites equipped with SAR sensors than optical sensors. Our approach in this paper is to leverage the availability and interpretability of high-resolution RGB satellite imagery. Our goal is to create a framework that can be readily deployed to perform hurricane damage assessment quickly in future events rather than depending on crowd-sourcing campaigns which could be time-consuming and rely on the availability and accuracy of volunteers. The framework only requires publicly available satellite imagery, such as the one used in this paper or aerial imagery collected from drones, and geographic information system (GIS) data for building coordinates and geolocation features.

\subsection{Deep learning with mixed data}
Recent studies have demonstrated that the information is much richer when combining images and other modality of data. In \cite{tang2015improving}, image classification can be improved by including location context, derived from Global Position System (GPS) tags, of the images. Similarly, \cite{kruk2019integrating} also successfully learn that images on Instagram and text caption can interact with each other to inform a more complex meaning that can explain their intent, contextual, and semiotic relationships. Perhaps most related to our work are the studies showing promising results in predicting housing price using traditional housing attributes, such as area, number of rooms, zipcode, etc. combined with the house interior/exterior photo \cite{ahmed2016house}, or the neighbourhood street and aerial views \cite{law2019take}. There is still a relatively smaller number of studies using multimodal data than those using images only. As highlighted by many authors in the field, there are various levels of challenges in collecting data and how to incorporate the non-image features effectively into the CNN model. In our work, we also encounter similar issues and data collection and preprocessing easily take up a major amount of work. Nonetheless, the result is really rewarding for us to achieve state-of-the-art performance in post-hurricane damage assessment. The computational cost, given the data is available, is still much more efficient than physical data collection and site survey. 

\section{Methodology}\label{sec:method_multimodal}
In this section, we describe our dataset and model architecture. 

\subsection{Data description}
\begin{figure}[ht!]
{\centering
\subfigure[\textit{`Undamaged Building'} using pre-event imagery]{\includegraphics[width=1.5in]{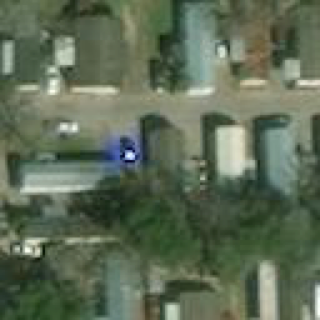}}
\subfigure[\textit{`Flooded/Damaged'} using post-event imagery]{\includegraphics[width=1.5in]{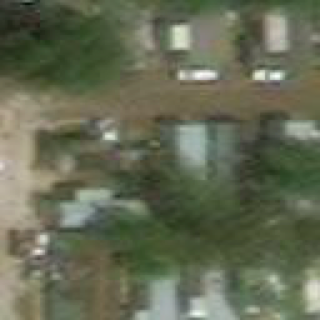}}
\subfigure[\textit{`Undamaged Building'} using post-event imagery]{\includegraphics[width=1.5in]{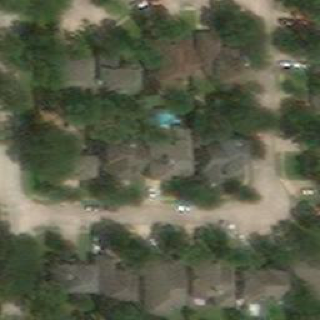}}
\subfigure[\textit{`Flooded/Damaged'} using post-event imagery]{\includegraphics[width=1.5in]{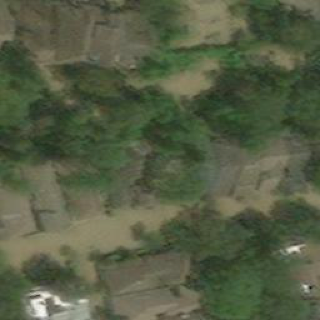}}
\\
\caption{\small{Two different ways to construct the dataset. The first way (a and b) uses different temporal information of the same location to label the data. The second way (c and d) uses different spatial information of the same timestamp to label the data.}}
\label{fig:data_construction}
}
\end{figure}

The data we used are the publicly available imagery captured after Hurricane Harvey event (post-event data), plus the coordinates annotated by crowd-sourcing campaign volunteers who identified whether a building is damaged or flooded, made available by DigitalGlobe \cite{tomnod}. The raw imagery data covering the Greater Houston area was captured in about four thousand strips ($\sim$400 million pixels ($\sim$1GB) with RGB bands per strip) in different days. In our previous work \cite{cao2020building}, we used the post-event imagery to crop the images at those coordinates to build the positive labels (\textit{Flooded/Damaged}), and pre-event data at the same coordinates to build the negative labels (\textit{Undamaged}). This approach using temporal difference to separate the data presents some limitations in terms of modelling and usability in the future. As can be seen in Figure~\ref{fig:data_construction}a and Figure~\ref{fig:data_construction}b, the positive and negative labels from different timestamps may have different color scale, hue, or saturation. In addition, there is flood water almost in every positive labels, which might lead the model into water detection rather than actual damage detection. 

In current work, we extract the data from the same post-event imagery (Figure~\ref{fig:data_construction}c and Figure~\ref{fig:data_construction}d, where it is inherently more difficult for the model to distinguish between damaged and undamaged building where flood water already invades most of the area. The color scale, hue, or saturation are also more consistent across the whole dataset to eliminate undesirable learning by the color scale.

\begin{figure*}[ht!]
{\begin{center}
\includegraphics[width=0.8\linewidth]{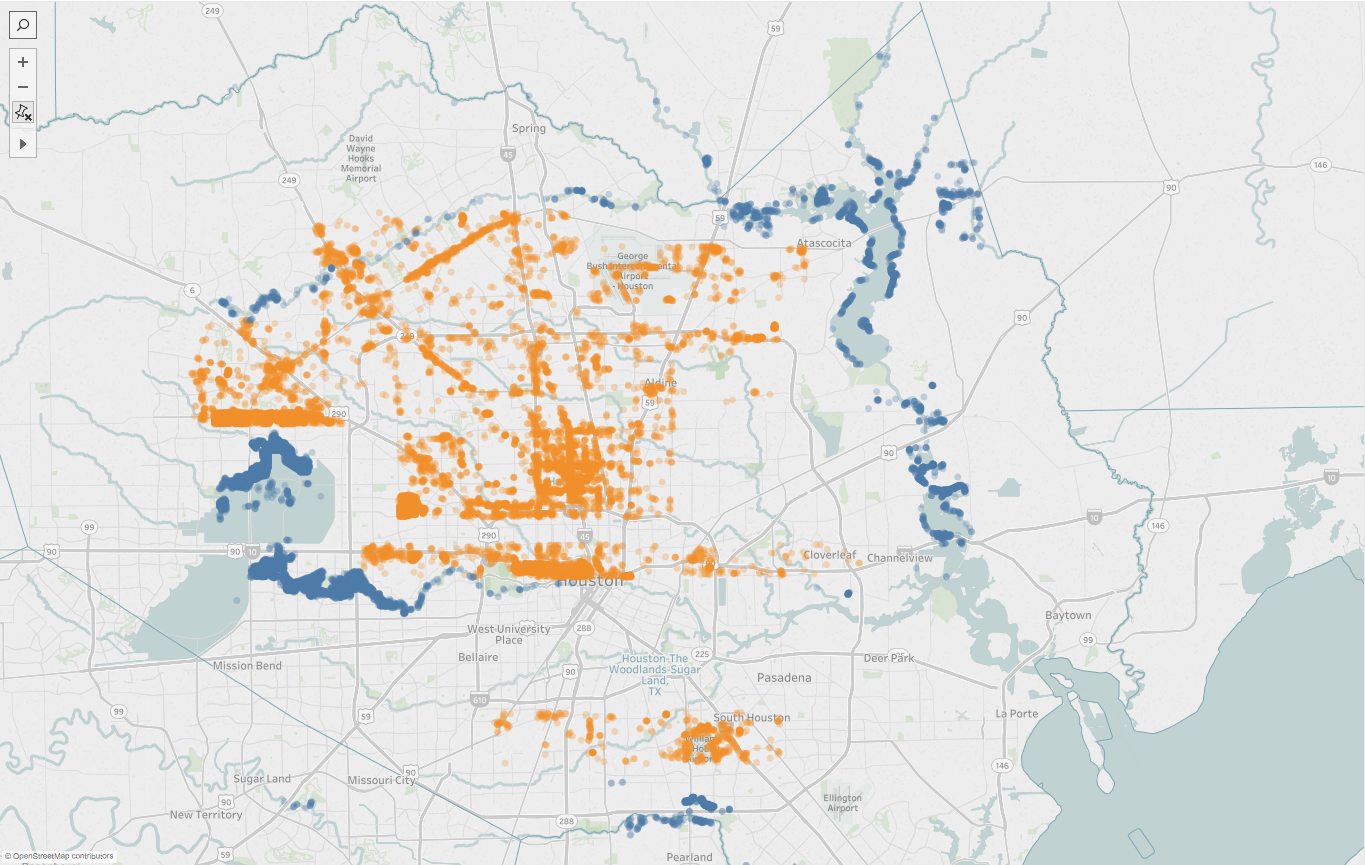}
\end{center}
\centering\caption{\small{\textit{Post-Hurricane Harvey `Flooded/Damaged'} and \textit{`Undamaged'}  building locations considered in this study.}}
\label{fig:houston-2types}
}
\end{figure*}

Since the coordinates provided by DigitalGlobe only include positive labels, we need to manually collect the negative labels ourselves as shown in Figure~\ref{fig:houston-2types}. From the OpenStreetMap API \cite{osm}, we divide the area into customized, much smaller strips to extract building coordinates that do not share the same footprint with the coordinates provided by DigitalGlobe's volunteers. To this end, we are assuming that the search by the volunteers are exhaustive, and every building coordinate not found by the volunteers is considered as undamaged. 

After collecting the set of coordinates for both labels, we use Google Map Developers API \cite{googlemap} to get the elevation at these coordinates to build the elevation feature for the dataset. From the same set of coordinates, we use QGIS GRASS API to find the distance from each coordinate to their nearest water body. The raster data for Texas area water bodies are provided by the United States Geological Survey (USGS) Geographic Information System Data \cite{usgs}.

\begin{figure*}[ht!]
{\centering
\subfigure[Simulation of 5-meter flood in Houston area]{\includegraphics[width=0.80\linewidth]{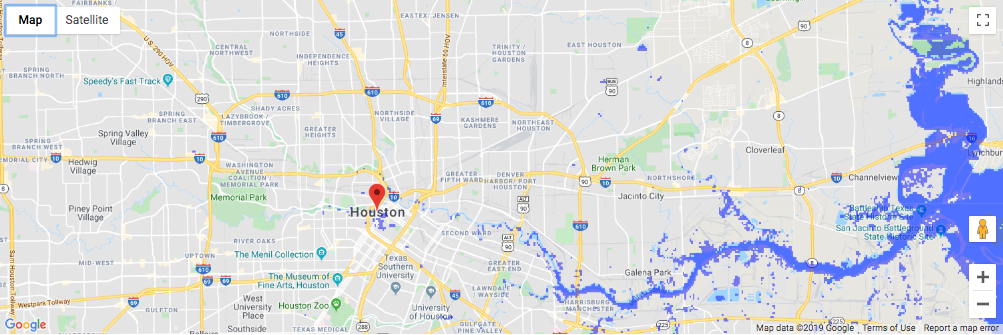}}
\subfigure[Simulation of 10-meter flood in Houston area]{\includegraphics[width=0.80\linewidth]{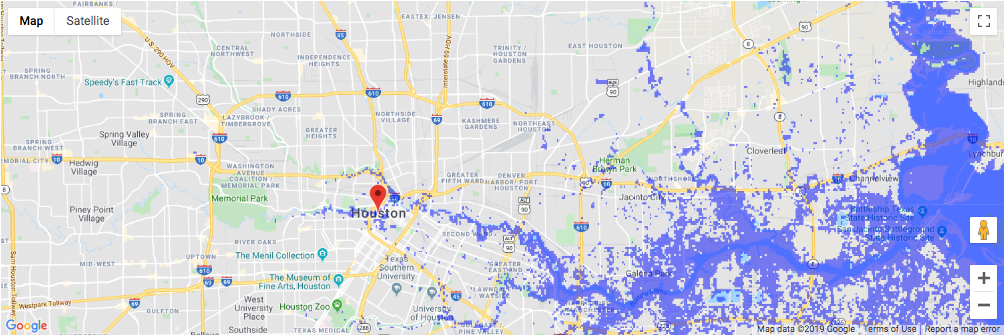}}
\subfigure[Simulation of 15-meter flood in Houston area]{\includegraphics[width=0.80\linewidth]{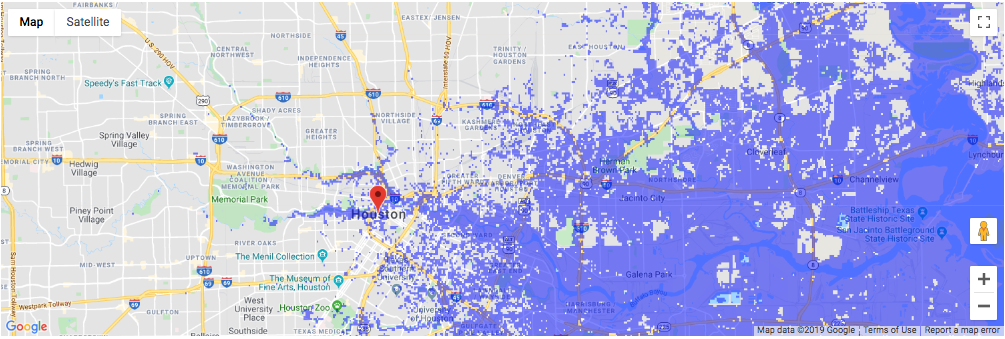}}
\\
\caption{\small{Simulation of different levels of flooding in the Houston area.}}
\label{fig:flood_simulation}
}
\end{figure*}

The rationale behind choosing distance from water bodies and elevation as extra geolocation features comes from some visualization and analyses. As can be seen in Figure~\ref{fig:houston-2types}, most of the \textit{`Flooded/Damaged'} buildings are very close to the major water bodies in the region. This is further confirmed through a flood map simulation \cite{floodmap} in the Houston area as shown in Figure~\ref{fig:flood_simulation}. As we increase the flood depth, areas around major water bodies have higher likelihood of being flooded.

To decide the second geolocation feature, we also make some strategic comparison between individual buildings' elevation levels and geographic coordinates. Initially, coordinates seem to be a logical choice to encode the neighbouring relationship of building cluster, which share similar disaster resilience characteristics and tend to be affected together. However, elevation can be even more informative. First, it can be used as an encoder for neighbouring houses as well, as nearby houses tend to not differ much in elevation. Second, we can capture an obvious physical behaviour, in which lower elevation may result in higher likelihood of getting flooded. Last but not least, elevation may be used to generalize to other regions, whereas coordinates practically cannot. Another region may have a different elevation, and different flood catchment capacity but the model is not expected to process absolute elevation value. We normalize the elevation to encode their relative difference within the region of interest. In future events, as long as their relative difference in elevation still prevail, we can still deploy the model trained using Houston data to quickly perform damage assessment over there.

\subsection{Model description}
The models presented in Section~\ref{sec:case_study_multimodal} are based on deep convolutional neural network. The model consists of an image encoder for the imagery, some fully connected layers to encode the geolocation features, some fusion layers to combine the two encoded information, and a class prediction layer. 

\begin{figure*}[!ht]
{\begin{center}
\includegraphics[width=0.8\linewidth]{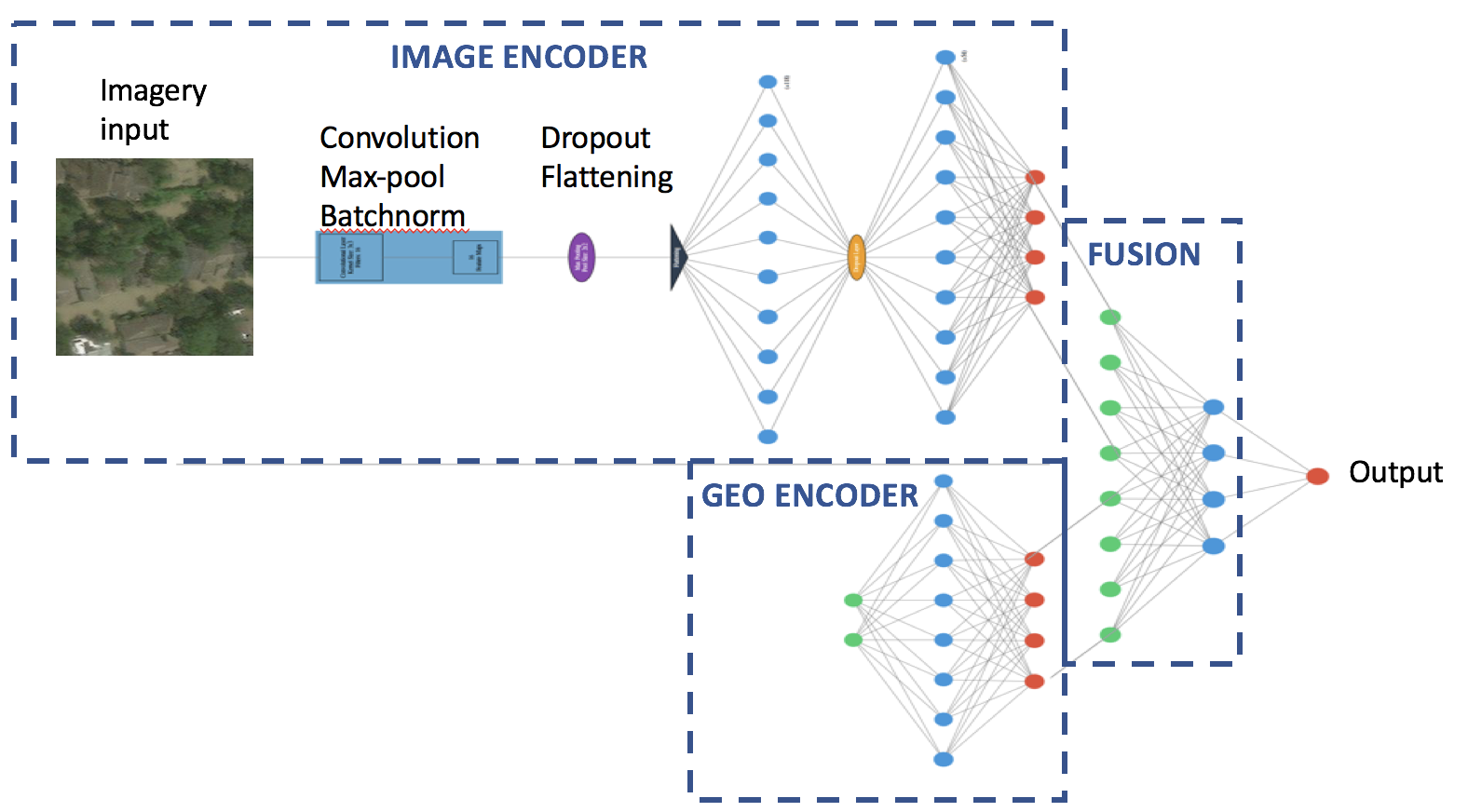} 
\end{center}
\centering\caption{\small{The mixed data neural network model that utilizes both satellite imagery and geolocation features to detect damaged building.}}
\label{fig:combined-python}
}
\end{figure*}

For image encoder, the same convolutional setup in \cite{cao2020building} is adopted with a sequence of convolution layers, max pooling layers, followed by a fully connected layer. At the end, the image encoder yields a 4-dimensional embedding for the imagery. For the geolocation encoder, two fully connected layers result in a 4-dimensional embedding. These two embeddings are concatenated to form a common embedding dimension of 8 in the fusion layers, which yield the final single node for class prediction after a few more fully connected layers (Figure~\ref{fig:combined-python}).

There are some hyper-parameter tuning works in the embedding size. We would like to investigate the effect of giving the same embedding sizes to the imagery and the geolocation features. From our previous work, we know that the image encoding works quite well with the image feature alone so there is no issue with using a small embedding size (e.g 4 dimensions). The question remains whether to give the geolocation the same or smaller embedding size since we start with only two features. This decision is informed through analyzing the performance of the model using purely imagery and geolocation features. As can be seen in Section~\ref{sec:case_study_multimodal}, geolocation features already provide a good signal to the likelihood of building damage, almost comparable to the imagery, which leads us to give the the equal embedding size in the combined model. 

\section{Implementation and Result}\label{sec:case_study_multimodal}
The network is trained to optimize with the Adam optimizer using the cross-entropy loss. We run all experiments on CentOS 7.7 (64-bit Linux) by training and validating our model using 5-fold cross-validation for 70 training epochs. In total, we spent 25 GPU hours to run all experiments. The model is built through the \textit{Keras} library with TensorFlow backend with a single NVIDIA K80 Tesla GPU with 64GB memory on a quad-core CPU machine.

The individual images are cropped from the original imagery at the window size of 256x256 due to its better performance in the mixed data model. We experimented with two different cropping window sizes, 128x128 and 256x256 pixels, as they yielded the highest performance metrics in our previous work using images only \cite{cao2020building}.  

This method relies heavily on the availability and quality of data and therefore poses some potential limitations. First, the imagery data was taken as a time series, with a lot of orthorectification and cloud coverage issues. It takes multiple iterations of visual inspection and manual processing to get a reasonable amount of usable data. Second, the geolocation data comes from different sources. Some data sources do not have complete data and/or use different formats. Preprocessing is intensive to join all the data together based on their geo-coordinates. Nevertheless, the entire process can be done computationally and saves substantial time and manpower required for traditional damage assessment practice such as post-event windshield survey. 

After cleaning and manual filtering, we are left with 13,993 positive samples (\textit{Damaged}) and 10,384 negative samples (\textit{Undamaged}) of unique coordinates. The dataset is split to have 67\% of the data as training data and 33\% as test data. The class distribution is preserved similarly to the original distribution in training and test data. The split is repeated 5 times to form 5 cross-validation sets, in each of which we train and test using the same model architecture to get both the mean and standard deviation of performance. 

We present our model performance results on the test data in Table~\ref{tab:experiment_multimodal} based on the probability threshold of 50\% to determine a class prediction. Due to class imbalance in our dataset, the metrics used here are accuracy, F1 score, precision, and recall. Between image feature and geolocation features, the former yields better precision. On the other hand, geolocation features seem to provide better recall than precision so it is more effective in detecting \textit{Damaged} samples. This is not surprising since the geolocation features are carefully designed and we expect building damage to follow physical laws. Generally, combining different types of data yields quite balanced performance and improves all metrics. Depending on priority of the model users, probability threshold can be adjusted to trade for more recall in order to identify more \textit{Damaged} samples. 

\begin{table*}[ht!]
\caption{Performance metrics across models}
    \begin{center}
    \begin{tabular}{|c||c|c|c|c|}
        \hline
        \multirow{2}{4em}{Method} & \multicolumn{4}{c|}{Metrics}
        \\
        & ACC & Precision & Recall & F1 score\\\hline\hline
        Img only & $79.5\pm8.3$\%&  $0.88\pm0.03$&  $0.64\pm0.30$&  $0.68\pm0.22$\\
        Geo only  & $88.6\pm1.4$\%&  $0.86\pm0.03$&  $0.97\pm0.003$&  $0.91\pm0.02$ \\
        \bf{Img + Geo}  &\boldmath{$97.47\pm2.5$\%} &  \boldmath{$0.91\pm0.14$}&  \boldmath{$0.99\pm0.003$}&  \boldmath{$0.94\pm0.08$}\\ \hline
    \end{tabular}
    \end{center}
\textit{Remarks}: Img only: model trained on image feature only; Geo only: model trained on geolocation features only; Img + Geo: model trained on both types of data. Each performance metric reported here shows the mean $\pm$ standard deviation across 5 cross-validation sets.
\label{tab:experiment_multimodal}
\end{table*}

\section{Conclusion}\label{sec:conclusion_multimodal}
We have demonstrated that damaged buildings can be detected with 97\%+ accuracy. Geolocation information can substantially improve the performance of CNN, and reduce the hyper-parameter tuning work. The model can be generalized to other regions and events as more data from more hazard events are aggregated. Since the geolocation features used are carefully chosen as relative elevation and relative proximity to water body, the model can be adapted to deploy to other regions and events without retraining. It could be that the relationship between elevation and flood likelihood is specific to regions, but we are only trying to capture the neighboring representation of the buildings through the similarity in their elevation. Therefore, to apply the model to another region, we might only need to adjust the class prediction threshold to gain more recall, if necessary. However, there is a trade-off between using too specific features to a hazard type (e.g. hurricane) such as the proximity to water bodies and generalization to other types (e.g earthquakes). It will be helpful for the disaster management community to train a few models validated on past events, that is ready for deployment for the next event. 

In future work, we hope to investigate incorporating other features such as catchment capacity or flood risk mapping to further improve the performance and robustness of the model. There are two other potential directions that can be pursued following our findings in this work. The first direction is rapid, real-time damage mapping of damaged buildings. From recent efficient and instant object detection algorithms such as \cite{redmon2016you, ren2015faster}, it would be possible to gather damage status of buildings through more accessible devices such as drones, freeing the reliance on satellite imagery. Since satellite imagery are considered to be complex to process due to their size and containing several objects at different scales \cite{sublime2017multi}, incorporating geolocation features to existing object detection algorithms can improve their metrics such as precision and recall. A potential challenge of this direction is the amount of labelled data required is usually quite large, which grows together with how complex the model is. This poses another issue of noisy or wrong labels, which leads to a second potential direction, label refinement. Recent studies have highlighted the needs of label refinement in the presence of noisy or wrong labels in large-scale datasets \cite{bagherinezhad2018label}, or in remote sensing data \cite{shang2020sar}. From our studies, geolocation features alone already inform substantial prior knowledge about the damage likelihood. We can use that information to correct the wrong labels as necessary.


%

\section*{Acknowledgment}
This work was partially supported by the National Science Foundation (NSF grant CMMI-1824681). We would like to thank Maxar for data sharing through their Open Data Program.

\newpage
\ifCLASSOPTIONcaptionsoff
  \newpage
\fi

\bibliographystyle{IEEEtran}
\bibliography{IEEEabrv,multimodal.bib}

\begin{IEEEbiography}[{\includegraphics[width=1.1in,height=1.25in,clip]{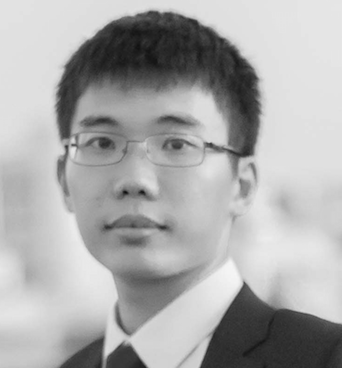}}]{Quoc Dung Cao}
is a Ph.D. student at the University of Washington - Seattle, where he is researching on developing computer vision and statistical applications to disaster management. His works involve analyzing satellite imagery and other GIS data to quantify damage level after a hurricane event. He is also investigating an infrastructure recovery trajectory estimation framework to aid in recovery planning for critical infrastructures.
\end{IEEEbiography}

\begin{IEEEbiography}[{\includegraphics[width=1.1in,height=1.25in,clip]{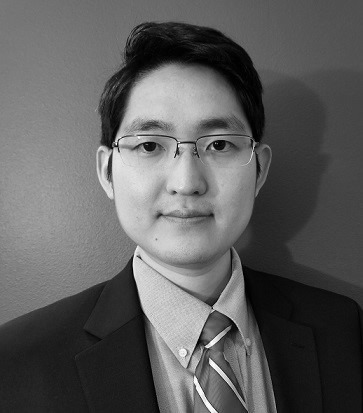}}]{Youngjun Choe}
received the B.Sc. degrees in Physics and Management Science from KAIST, Korea in 2010, and both the M.A. degree in Statistics and the Ph.D. degree in Industrial \& Operations Engineering from the University of Michigan, Ann Arbor, MI, USA in 2016. 

He is currently an Assistant Professor of Industrial \& Systems Engineering at the University of Washington, Seattle, WA, USA. His research centers around developing statistical methods to study extreme events using empirical and simulated data. 
\end{IEEEbiography}





\end{document}